\documentclass[10pt,twocolumn,letterpaper]{article}

\usepackage{cvpr}              
\usepackage{graphicx}     
\usepackage{subcaption}   
\usepackage{multirow}
\usepackage{tabularx}
\usepackage{booktabs}
\usepackage{colortbl}
\usepackage{amsmath}

\usepackage[symbol]{footmisc} 

\definecolor{cvprblue}{rgb}{0.21,0.49,0.74}
\usepackage[pagebackref,breaklinks,colorlinks,allcolors=cvprblue]{hyperref}

\def\paperID{2382} 
\def\confName{CVPR}
\def\confYear{2025}

\title{Timestep Embedding Tells: It's Time to Cache for Video Diffusion Model}

\author{Feng Liu\textsuperscript{1}\thanks{Work was done during internship at Alibaba Group.}
\quad 
Shiwei Zhang\textsuperscript{2}\thanks{Project Leader.}\quad
Xiaofeng Wang\textsuperscript{1,3}\quad
Yujie Wei\textsuperscript{4}\quad
Haonan Qiu\textsuperscript{5}\\
Yuzhong Zhao\textsuperscript{1}\quad
Yingya Zhang\textsuperscript{2}\quad
Qixiang Ye\textsuperscript{1}\quad
Fang Wan\textsuperscript{1}\thanks{Corresponding author.}\\
\textsuperscript{1}University of Chinese Academy of Sciences
\quad
\textsuperscript{2}Alibaba Group\\
\textsuperscript{3}Institute of Automation, Chinese Academy of Sciences\\
\textsuperscript{4}Fudan University\quad
\textsuperscript{5}Nanyang Technological University\\
Project Page: {\color{magenta}https://liewfeng.github.io/TeaCache}
}

\begin{document}

\twocolumn[{%
\renewcommand\twocolumn[1][]{#1}%
\maketitle
\vspace{-0.7cm}
\includegraphics[width=\linewidth]{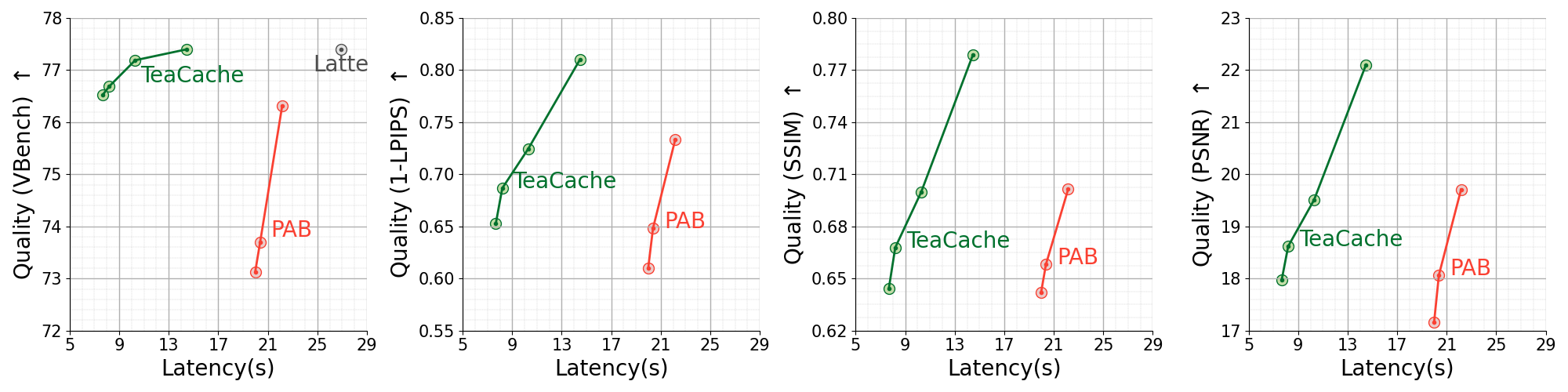}
}
\vspace{-0.2cm}
\captionof{figure}{\textbf{Quality-latency comparison of video diffusion models.} Visual quality versus latency curves of the proposed TeaCache approach and PAB~\cite{zhao2024real} using Latte~\cite{ma2024latte}.
TeaCache significantly outperforms PAB in both visual quality and efficiency. Latency is evaluated on a single A800 GPU for 16-frame video generation under 512 $\times$ 512 resolution.
\vspace{2em}
\label{fig:shot}
}
]

\footnotetext{\textsuperscript{*}Work was done during internship at Alibaba Group.}
\footnotetext{\textsuperscript{$\dagger$}Project Leader. \textsuperscript{$\ddagger$}Corresponding author.}

\begin{abstract} 
As a fundamental backbone for video generation, diffusion models are challenged by low inference speed due to the sequential nature of denoising.
Previous methods speed up the models by caching and reusing model outputs at uniformly selected timesteps.
However, such a strategy neglects the fact that differences among model outputs are not uniform across timesteps, which hinders selecting the appropriate model outputs to cache, leading to a poor balance between inference efficiency and visual quality.
In this study, we introduce Timestep Embedding Aware Cache (TeaCache), a training-free caching approach that estimates and leverages the fluctuating differences among model outputs across timesteps. 
Rather than directly using the time-consuming model outputs, TeaCache focuses on model inputs, which have a strong correlation with the modeloutputs while incurring negligible computational cost.
TeaCache first modulates the noisy inputs using the timestep embeddings to ensure their differences better approximating those of model outputs. 
TeaCache then introduces a rescaling strategy to refine the estimated differences and utilizes them to indicate output caching.

Experiments show that TeaCache achieves up to 4.41$\times$ acceleration over Open-Sora-Plan with negligible (-0.07\% Vbench score) degradation of visual quality. 
\end{abstract}    

\begin{figure*}[t]
  \centering
    \includegraphics[width=1.0\linewidth]{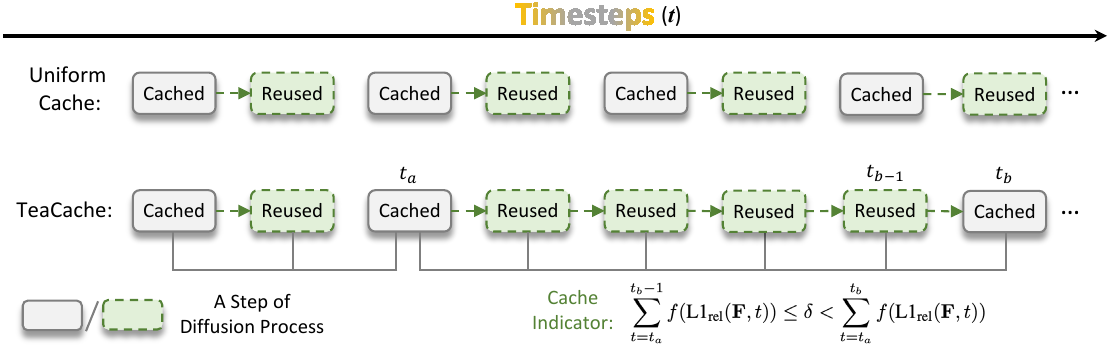}
  \caption{Comparison of the proposed TeaCache and the conventional uniform caching strategy for DiT models during inference. TeaCache is capable of selectively caching informative intermediate model outputs during the inference process, and therefore accelerates the DiT models while maintaining its performance. $\textbf{F}$ and $t$ respectively denote the model inputs of noisy input and timestep embedding. $L1_{rel}$ and $f$ are difference estimation functions of model inputs. $\delta$ is an indicator threshold of whether to cache a model output or not.}
  \label{fig:method_difference}
\end{figure*}

\vspace{-0.5cm}
\section{Introduction}
\label{sec:intro}

Recent years have witnessed the emergence of diffusion models~\cite{dhariwal2021diffusion, ho2020denoising, sohl2015deep, song2019generative}, as a fundamental backbone for visual generation.
The model architecture has evolved from U-Net ~\cite{ramesh2022hierarchical, saharia2022photorealistic, blattmann2023stable} to diffusion transformers (DiT)~\cite{peebles2023scalable}, which greatly increased model capacities.
Empowered by DiT, video generation models~\cite{Open-Sora, Open-Sora-Plan, ma2024latte, yang2024cogvideox, Vchitect, Mochi} 
have reached a groundbreaking level.

Despite of the substantial efficacy of these powerful models, their inference speed remains a pivotal impediment to wider adoption~\cite{li2024snapfusion}. This core limitation arises from the sequential denoising procedure inherent to their reverse phase, which inhibits parallel decoding~\cite{shih2024parallel}. Moreover, as model parameters scale up and the requirements for higher resolution and longer durations of videos escalate~\cite{chen2024pixart, yang2024cogvideox}, the inference process experiences a further decline in speed.

To accelerate the visual generation procedure,
distillation~\cite{sauer2023adversarial, wang2023videolcm, meng2023distillation} and post-training~\cite{chen2024q, ma2024learning} are employed. 
However, these methods typically require extra training, which implies substantial computational cost and data resources.
An alternative technological pathway is to leverage the caching mechanism~\cite{smith1982cache, goodman1983using, albonesi1999selective}, which does not require additional training to maintain the performance of diffusion models.
These methods~\cite{xu2018deepcache, selvaraju2024fora, zhao2024real, chen2024delta} find that the model outputs are similar between the consecutive timesteps when denoising and propose to reduce redundancy by caching model outputs in a uniform way, Fig.~\ref{fig:method_difference}(upper).
Nevertheless, when the output difference between consecutive timesteps varies, the uniform caching strategy lacks flexibility to maximize the cache utilization.

In this study, we aim to develop an novel caching approach by fully utilizing the fluctuating differences among outputs of the diffusion model across timesteps.
The primary challenge is: when can we reuse cached output to substitute the current timestep's output? Intuitively, this is possible when the current output is similar with the cached output, Fig.~\ref{fig:method_difference}(upper).
Unfortunately, such difference is not predictable before the current output is computed. Consequently, without the guidance of difference, the uniformly cached outputs becomes redundant and the inference efficiency remains low.

To conquer this challenge, we propose Timestep Embedding Aware Cache (TeaCache), a training-free caching strategy. TeaCache leverages the following prior: {\textbf{\textit{There exists a strong correlation between a model's inputs and outputs.}} If a transformation relationship can be established between the input and output difference, one can utilizes the difference among inputs as an indicator of whether the corresponding outputs need to be cached, Fig.~\ref{fig:method_difference}(lower). 
Since inputs are readily accessible, this approach would significantly reduce computation cost.
We then delve into the inputs of diffusion models: a noisy input, a timestep embedding, and a text embedding. The text embedding remains constant throughout the denoising process and cannot be used to measure the difference of input across timesteps.
As for the timestep embedding, it changes as timesteps progress but is independent of the noisy input and text embedding, making it difficult to fully reflect the input information. 
The noisy input, on the other hand, is gradually updated during the denoising process and contains information from the text embedding, but it is not sensitive to timesteps. 
To accurately describe the model inputs and ensure their strong correlation with the outputs, TeaCache follows the inference process of diffusion and employ the timestep-embedding modulated noisy input
as the final input embeddings, among which the difference are then used to estimated the output difference.

It is noteworthy that the input difference estimated above still exhibits a scaling bias relative to the output difference, which has been observed through empirical studies. That is because this strategy only captures the correlation trend between input difference and output difference. Considering that both input and output differences are already scalars, TeaCache further introduces a simple polynomial fitting procedure to estimate the scaling factors between them. 
With the correction of the scaling factors, the input difference can accurately reflect the output difference and is ultimately used as an indicator of whether the outputs need to be cached, Fig.~\ref{fig:method_difference}(lower).

The contributions of this paper include:
\begin{itemize}[leftmargin=*]
    \item We propose TeaCache, a training-free approach which is completely compatible with DiT diffusion models, to estimate the difference of model outputs, selectively cache model outputs and speed up the inference process.
    \item We propose a simple-yet-effective two-stage strategy to estimate the difference of model output through model input. The proposed strategy uses timestep-embedding modulated noisy input to perform coarse estimation and a polynomial fitting procedure for refinement.

    \item TeaCache speeds up SOTA generation models, Open-Sora \cite{Open-Sora}, Open-Sora-Plan \cite{Open-Sora-Plan}, and Latte \cite{ma2024latte}, (PAB~\cite{zhao2024real}) with large margins at negligible quality cost, Fig.~\ref{fig:shot}.
\end{itemize}

\section{Related Work}
\subsection{Diffusion Model}
In the realm of generative models, diffusion models \cite{ho2020denoising, sohl2015deep} have become foundational due to their exceptional ability to produce high-quality and diverse outputs. Initially developed with the U-Net architecture, these models have demonstrated impressive performance in image and video generation \cite{ramesh2022hierarchical, rombach2022high, ho2022video, saharia2022photorealistic, wei2024dreamvideo, wei2024dreamvideo2, wang2023modelscope, chen2023videocrafter1, chen2024videocrafter2}. 

However, the scalability of U-Net-based diffusion models is inherently constrained, posing challenges for applications requiring larger model capacities for enhanced performance. To address this limitation, Diffusion transformers (DiT) \cite{peebles2023scalable} represent a significant advancement. By utilizing the scalable architecture of transformers \cite{vaswani2017attention}, DiT provides an effective means to increase model capacity.
A notable achievement in this field is the advancement in generating long videos through the large-scale training of Sora \cite{Sora}, which employs a transformer-based Diffusion architecture for comprehensive simulations of the physical world. This underscores the considerable impact of scaling transformer-based Diffusion models.
An increasing number of studies have adopted the Diffusion transformer as the noise estimation network~\cite{chen2023pixart, chen2024pixart, Open-Sora, Open-Sora-Plan, ma2024latte, yang2024cogvideox}.

\subsection{Diffusion Model Acceleration}
Despite the notable performance of Diffusion models in image and video synthesis, their significant inference costs hinder practical applications. Efforts to accelerate Diffusion model inference fall into two primary categories. First, techniques such as DDIM~\cite{song2020denoising} allow for fewer sampling steps without sacrificing quality. Additional research has focused on efficient ODE or SDE solvers~\cite{song2019generative, jolicoeur2021gotta, lu2022dpm, karras2022elucidating, lu2022dpm++}, using pseudo numerical methods for faster sampling. Second, approaches include distillation~\cite{salimans2022progressive, wang2023videolcm}, quantization~\cite{li2024q, he2024ptqd, so2024temporal, shang2023post}, and distributed inference~\cite{li2024distrifusion} are employed to reduce the workload and inference time.  

However, these methods often demand additional resources for fine-tuning or optimization. Some training-free approaches~\cite{bolya2023token, wang2024attention} streamline the sampling process by reducing input tokens, thereby eliminating redundancy in image synthesis. Other methods reuse intermediate features between successive timesteps to avoid redundant computations~\cite{wimbauer2024cache, so2023frdiff, zhang2024cross}. DeepCache~\cite{xu2018deepcache} and Faster Diffusion~\cite{li2023faster} utilize feature caching to modify the UNet Diffusion, thus enhancing acceleration. FORA~\cite{selvaraju2024fora} and $\triangle$-DiT~\cite{chen2024delta} adapts this mechanism to DiT by caching residuals between attention layers. PAB~\cite{zhao2024real} caches and broadcasts intermediate features at various timestep intervals based on different attention block characteristics for video synthesis. 
AdaCache~\cite{kahatapitiya2024adaptive} proposes to dynamically adjusts feature caching strategies based on content complexity. 
FasterCache~\cite{lv2024fastercache} observes significant redundancy in CFG and optimizes the reuse of conditional and unconditional outputs.
While these methods have improved Diffusion efficiency, enhancements for DiT in visual synthesis remain limited.
\begin{figure*}[t]
    \centering
    \begin{minipage}{0.8\textwidth}
    \centering
    \begin{subfigure}{0.3\textwidth}
        \centering
        \includegraphics[width=\textwidth]{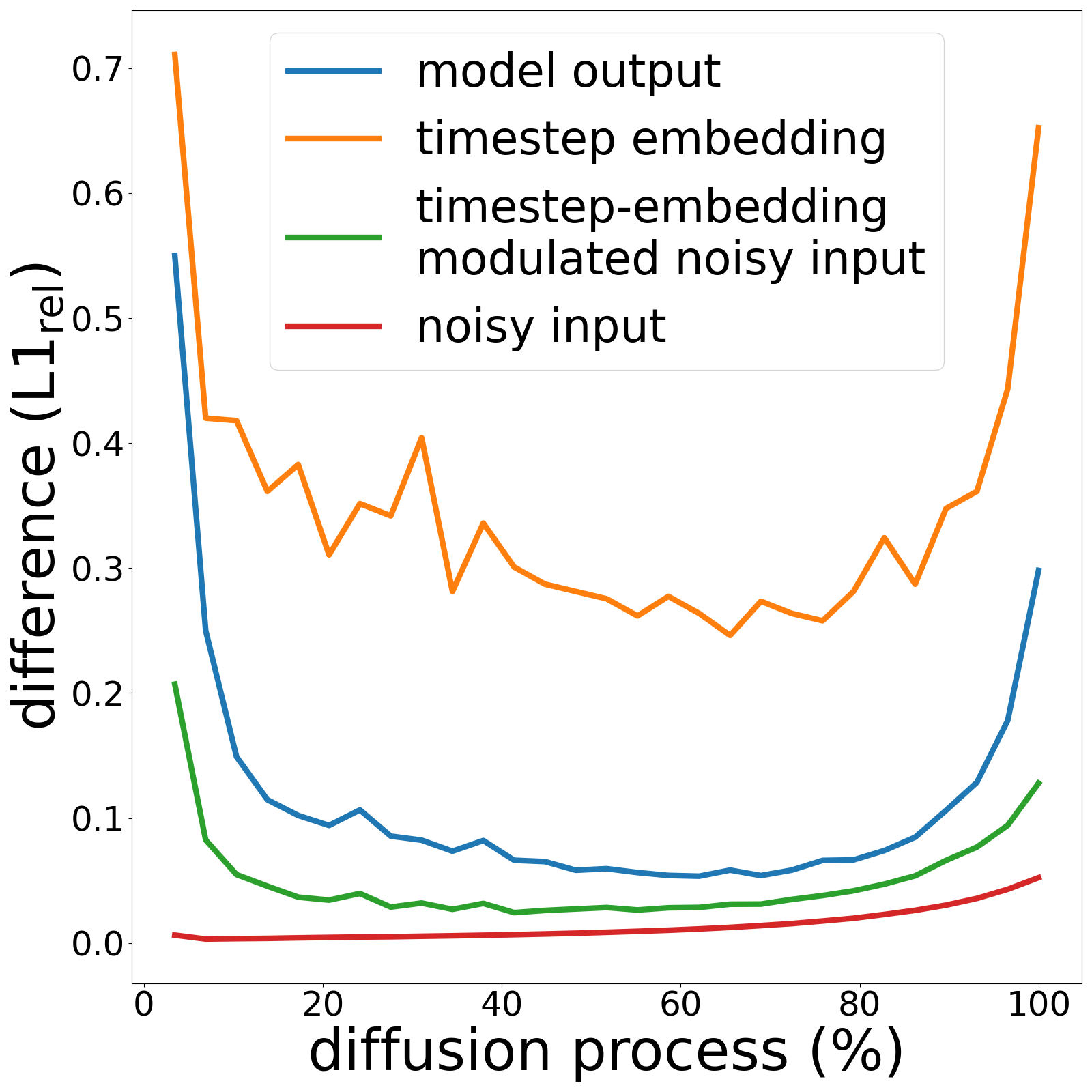} 
        \caption{Open Sora}
    \end{subfigure}
    \hfill
    \begin{subfigure}{0.3\textwidth}
        \centering
        \includegraphics[width=\textwidth]{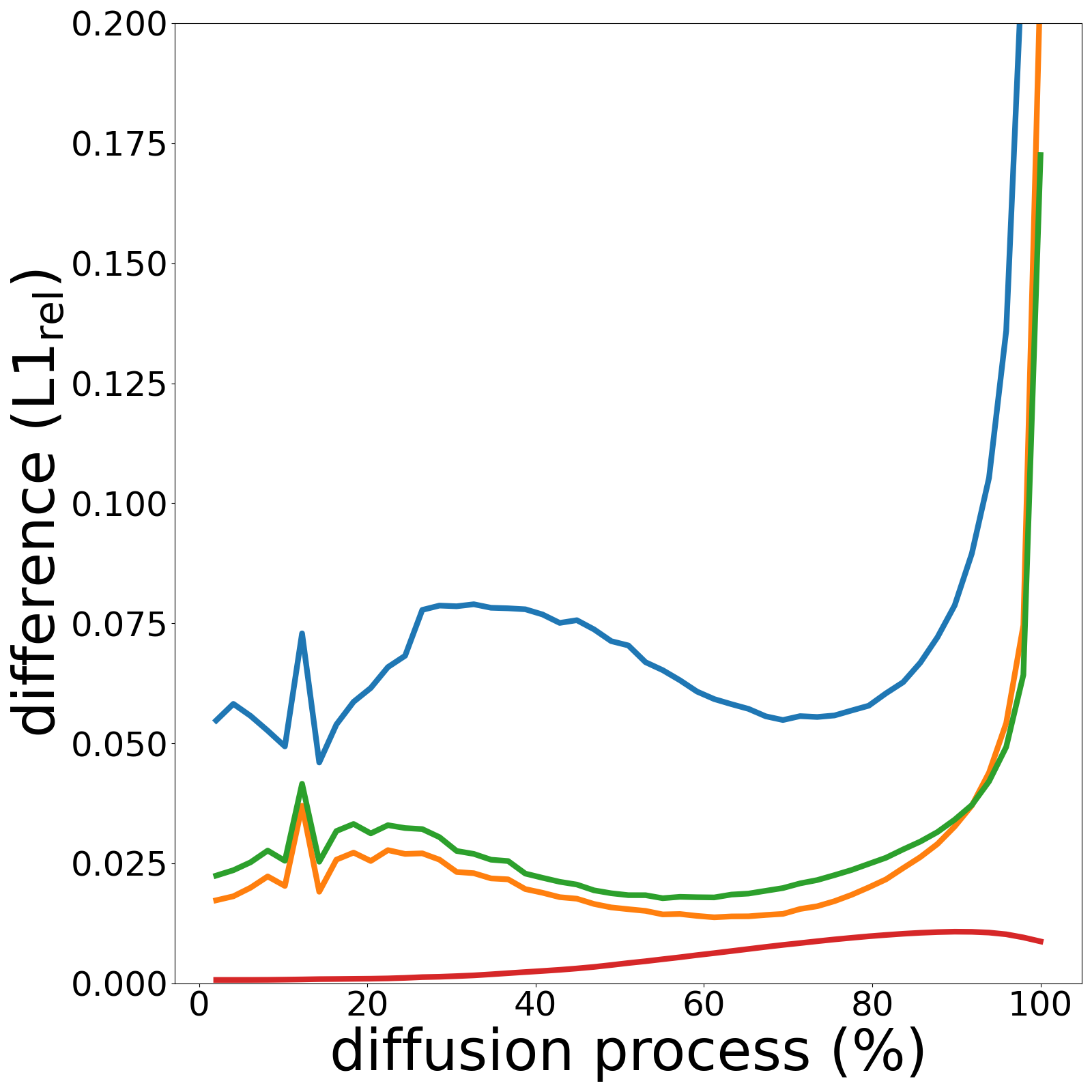} 
        \caption{Latte}
    \end{subfigure}
    \hfill
    \begin{subfigure}{0.3\textwidth}
        \centering
        \includegraphics[width=\textwidth]{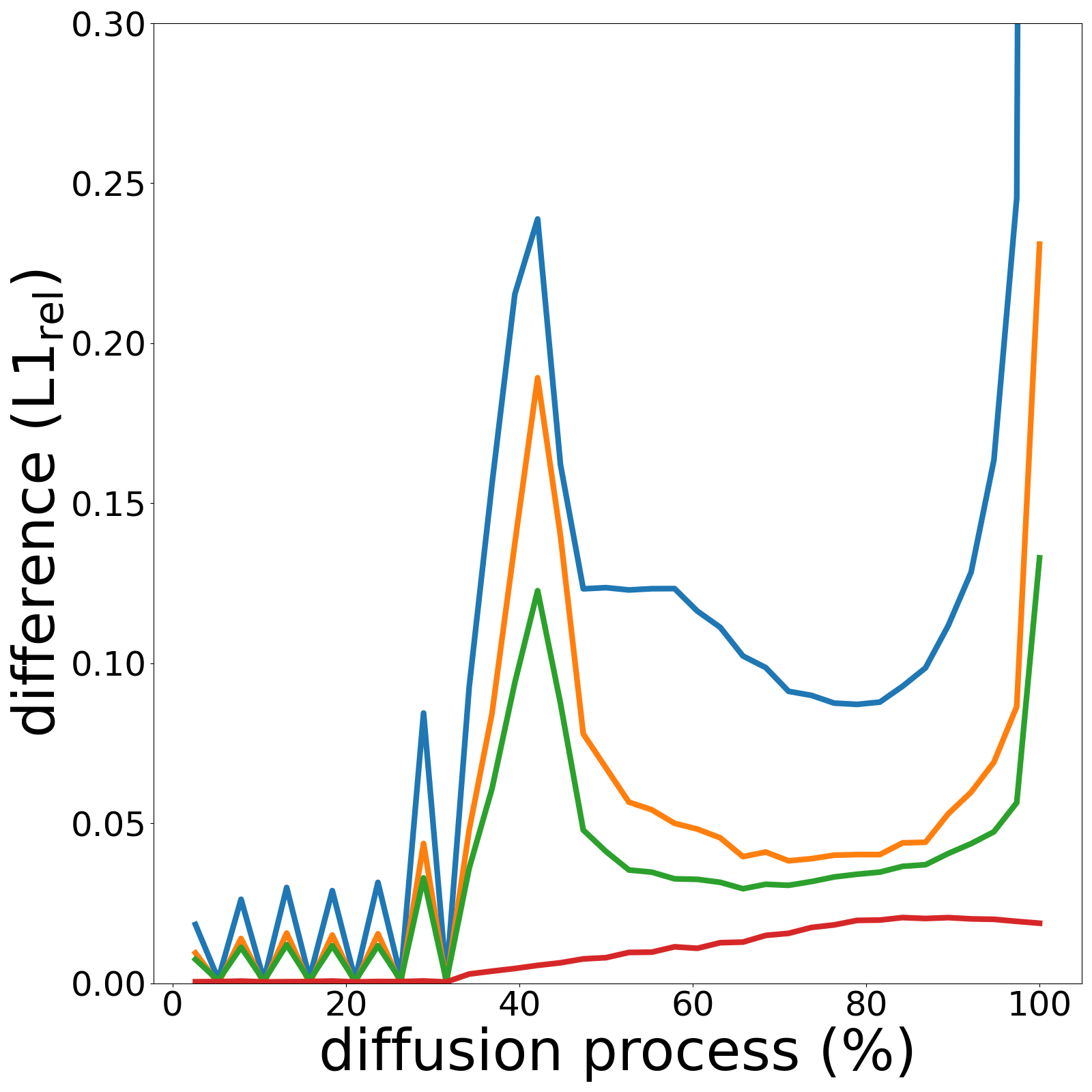} 
        \caption{OpenSora-Plan}
    \end{subfigure}
    \end{minipage}
    \vspace{-0.2cm}
    \caption{Visualization of input differences and output differences in consecutive timesteps of Open Sora, Latte, and OpenSora-Plan.
    Timestep embedding and Timestep embedding modulated noisy input have strong correlation with model output.}
    \label{fig:difference}
\end{figure*}

\section{Methodology}

\subsection{Preliminaries}
\textbf{Denoising Diffusion Models.} Diffusion models simulate visual generation through a sequence of iterative denoising steps. The core idea is to start with random noise and progressively refine it until it approximates a sample from the target distribution. During the forward diffusion process, Gaussian noise is incrementally added over T steps to a data point $\mathbf{x}_{0}$ sampled from the real distribution $q(\mathbf{x})$:
\begin{equation}
\mathbf{x}_{t}=\sqrt{\alpha_{t}} \mathbf{x}_{t-1}+\sqrt{1-\alpha_{t}} \mathbf{z}_{t} \quad \text { for } \quad t=1, \ldots, T
\end{equation}
where $\alpha_t \in [0,1]$ governs the noise level, and $\mathbf{z}_t \sim \mathcal{N}(\mathbf{0}, \mathbf{I})$ represents Gaussian noise. As $t$ increases, $\mathbf{x}_t$ becomes progressively noisier, ultimately resembling a normal distribution $\mathcal{N}(\mathbf{0}, \mathbf{I})$ when $t=T$. The reverse diffusion process is designed to reconstruct the original data from its noisy counterpart:
\begin{equation}
\label{eq:update}
p_\theta(\mathbf{x}_{t-1} \mid \mathbf{x}_t) = \mathcal{N}(\mathbf{x}_{t-1}; \mu_\theta(\mathbf{x}_t, t), \Sigma_\theta(\mathbf{x}_t, t)),
\end{equation}
where $\mu_\theta$ and $\Sigma_\theta$ are learned parameters defining the mean and covariance.

\begin{figure}[t]
  \centering
    \includegraphics[width=1.0\linewidth]{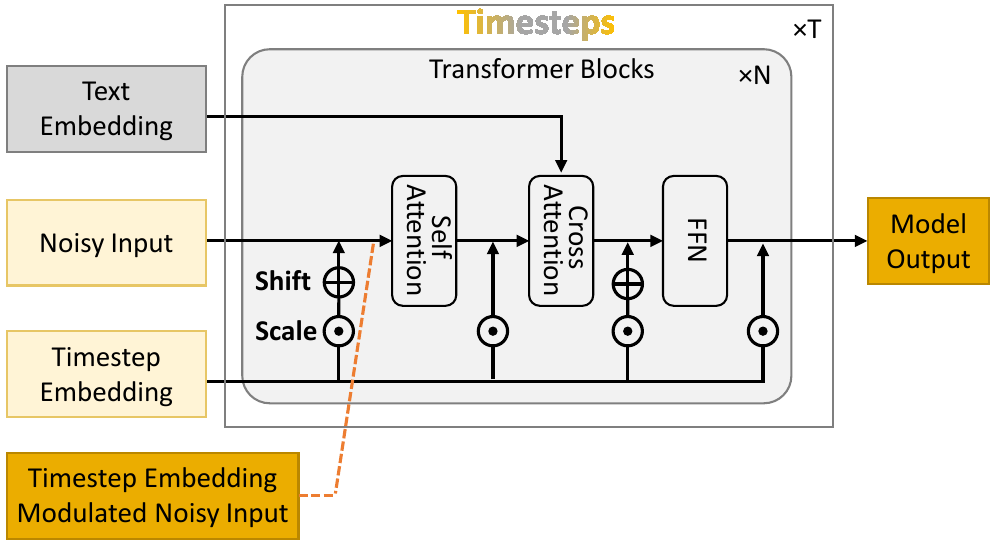}
  \caption{Diffusion module of the visual generation model with transformer. Normalization layer is omitted for simplicity. Timestep embedding modulates the magnitude of block input and output thus has the potential to indicate the variation of output. The output is updated with Eq.~\ref{eq:update} at different timesteps.}
  \label{fig:block}
\end{figure}

\textbf{Timestep Embedding in Diffusion Models.} 
The diffusion procedures are usually splitted to one thousand timesteps during training phase and dozens of timesteps during inference phase. Timestep defines the strength of noise to be added or removed in the diffusion procedures, which is an important input of the diffusion model. Specifically, the scalar timestep $t$ is firstly transformed to timestep embedding through sinusoidal embedding and multilayer perception module:
\begin{equation}
\mathbf{T}_{t} = MLP(sinusoidal(t))\quad \text { for } \quad t=1, \ldots, T.
\end{equation}
Timestep embedding then modulates the input and output of the Self Attention Layer and Feed Forward Network (FFN) in each Transformer block, as shown in Fig.\ref{fig:block}. Thus, timestep embedding can significantly affect the magnitude of the model output.

\begin{figure*}[t]
    \centering
    \begin{minipage}{0.8\textwidth}
    \centering
    \begin{subfigure}{0.3\textwidth}
        \centering
        \includegraphics[width=\textwidth]{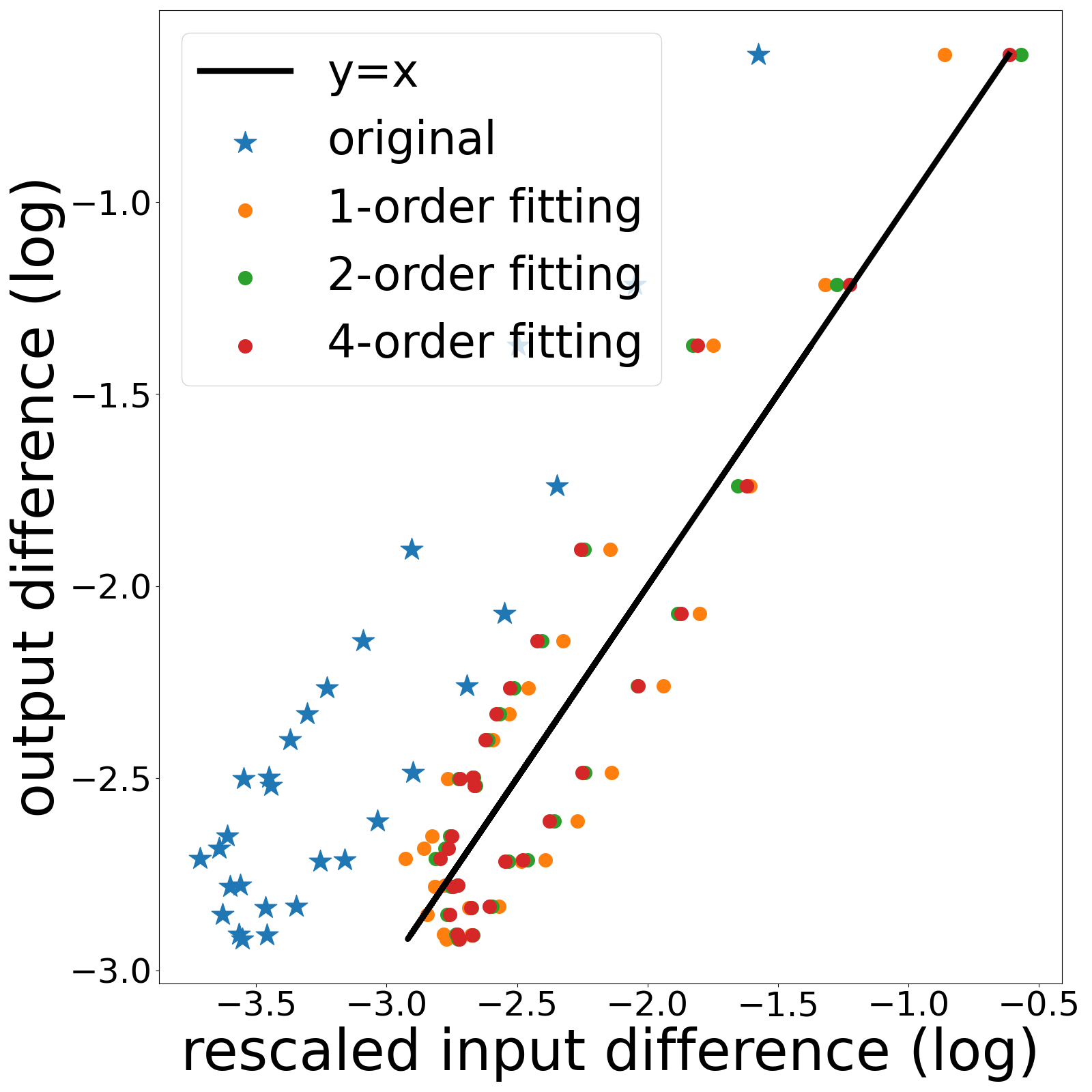} 
        \caption{Open Sora}
    \end{subfigure}
    \hfill
    \begin{subfigure}{0.3\textwidth}
        \centering
        \includegraphics[width=\textwidth]{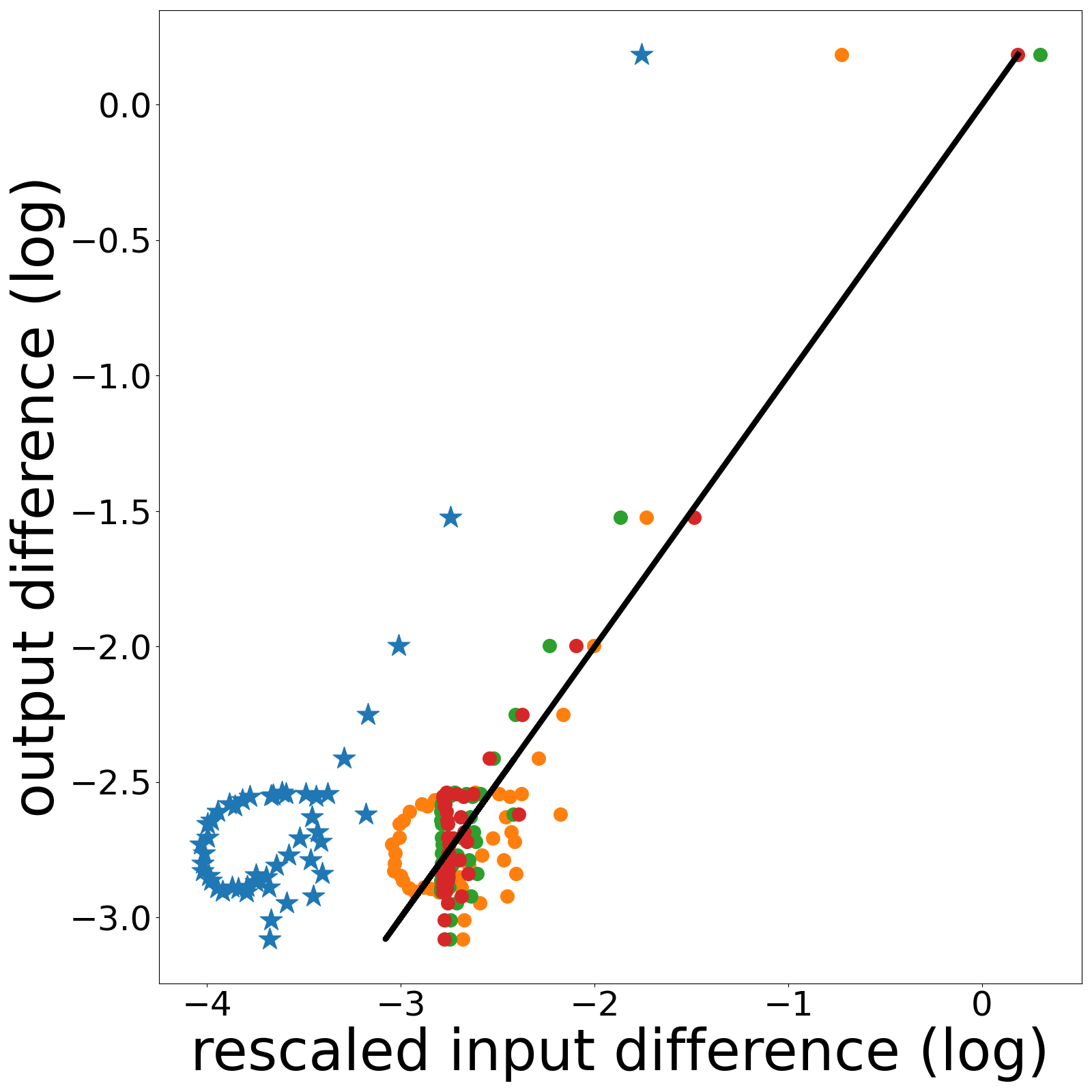} 
        \caption{Latte}
    \end{subfigure}
    \hfill
    \begin{subfigure}{0.3\textwidth}
        \centering
        \includegraphics[width=\textwidth]{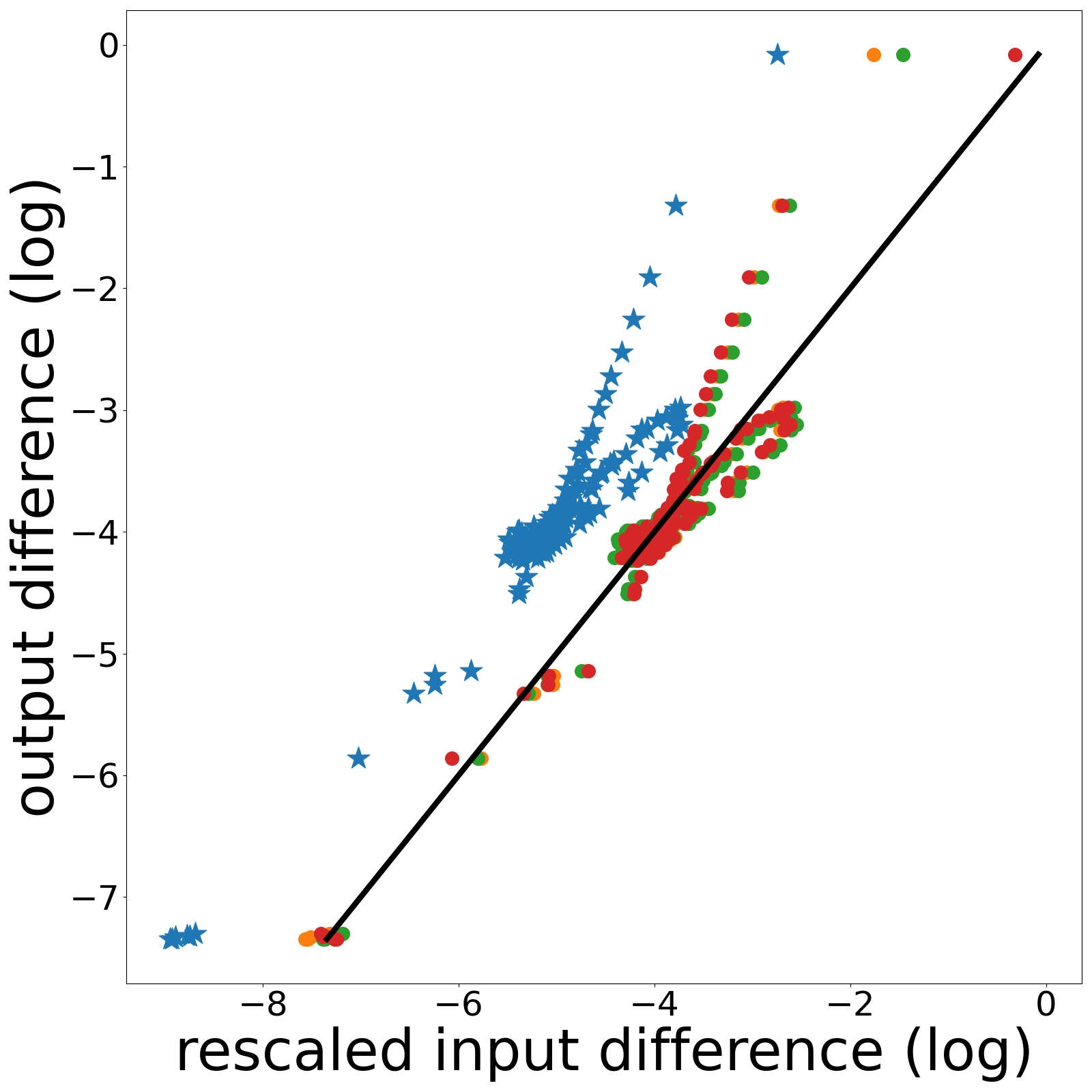}
        \caption{OpenSora-Plan}
    \end{subfigure}
    \end{minipage}
    \vspace{-0.2cm}
    \caption{Visualization of corelation of input differences and output differences in consecutive timesteps of Open Sora, Latte, and OpenSora-Plan. 
    The original data points deviate a lot from the linear corelation. Polynomial fitting reduces the gap.}
    \label{fig:fitting}
\end{figure*}

\begin{figure}[t]
  \centering
    \includegraphics[width=1.0\linewidth]{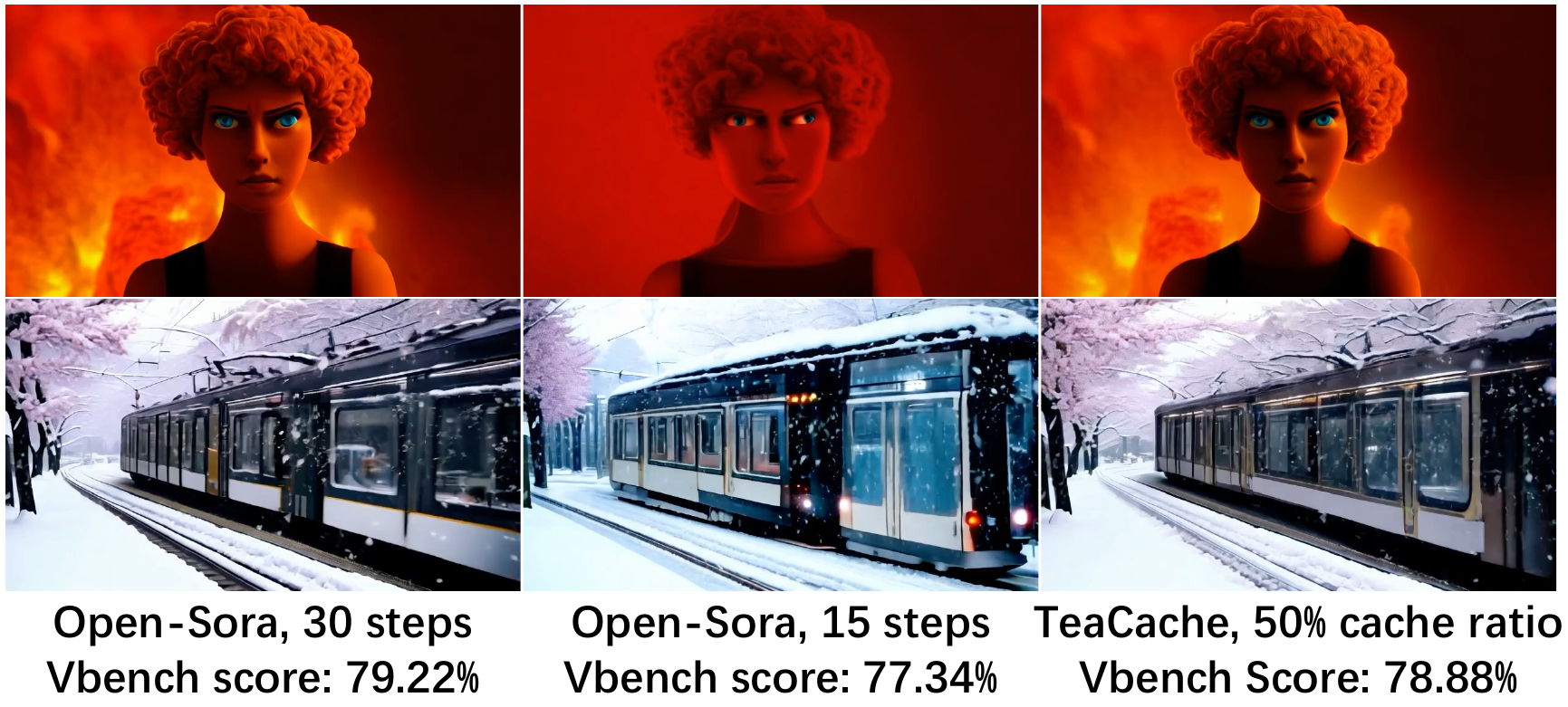}
  \caption{Caching Mechanism \textit{v.s.} Reducing Timesteps. Reducing inference timesteps suffers from deteriorated visual quality while TeaCache maintains visual quality.}
  \label{fig:few timestep}
\end{figure}

\subsection{Analysis}
To investigate the correlation between model output and input, we perform an in-depth analysis of their behaviors during the diffusion process. 

\textbf{Model outputs:} Ideally, if we could obtain the model outputs in advance, we could directly measure the difference between outputs at adjacent timesteps and decide whether to cache the outputs based on their difference. 
Following ~\cite{wimbauer2024cache}, we use the relative L1 distance as our metric. For instance, the relative L1 distance $\text{L1}_{\text{rel}}(\mathbf{O}, t)$ for output embedding $\mathbf{O}_t$ at timestep $t$ is calculated as follows:
\begin{equation}
\text{L1}_{\text{rel}}(\mathbf{O}, t) = \frac{\|\mathbf{O}_t - \mathbf{O}_{t+1}\|_1}{\|\mathbf{O}_{t+1}\|_1}
\label{eq:difference_measurement}
\end{equation}
where a large $\text{L1}_{\text{rel}}(\mathbf{O}, t)$ indicates that $\mathbf{O}_t$ is informative relative to $\mathbf{O}_{t+1}$ and should be cached; otherwise, a small $\text{L1}_{\text{rel}}(\mathbf{O}, t)$ indicates that $\mathbf{O}_{t+1}$ and $\mathbf{O}_t$ are similar to each other and therefore $\mathbf{O}_{t+1}$ could be reused to replace $\mathbf{O}_{t}$. Therefore, Eq.~\ref{eq:difference_measurement} can be used to define a criterion for determining whether the model outputs should be cached.

However, in most cases, the model outputs cannot be obtained in advance, making the above approach infeasible. To address this issue, an intuitive idea is that if we can efficiently estimate the difference of the model outputs, we can leverage it to design a caching strategy. Fortunately, it is well-known that the model inputs and outputs are strongly correlated. Based on this insight, we analyzed the model inputs and conducted detailed experiments to investigate their correlation with the model outputs.

\textbf{Model inputs:} We consider the inputs of diffusion model: text embedding, timestep embedding, and noisy input, as shown in Fig.~\ref{fig:block}. Since the text embedding remains constant throughout the diffusion process, it can't be used to measure the difference of inputs across timestep. Therefore, text embedding is excluded from analysis. As for the timestep embedding, it changes as timesteps progress but is independent of the noisy input and text embedding, making it difficult to fully reflect the information of the input. The noisy input, on the other hand, is gradually updated during the denoising process and contains information from the text embedding, but it is not sensitive to timesteps. To comprehensively represent the model inputs and ensure their correlation with the outputs, we ultimately utilized the timestep embedding modulated noisy input at the Transformer’s input stage as the final input embedding, as illustrated in the Fig.~\ref{fig:block}.

\textbf{Experimental analysis:}
To derive a robust conclusion, we make analysis using the metric defined in Eq.~\ref{eq:difference_measurement} to compute the difference of model inputs and outputs on three distinct video generation models: Open Sora~\cite{Open-Sora}, Latte~\cite{ma2024latte}, and OpenSora Plan~\cite{Open-Sora-Plan}. As illustrated in Fig.~\ref{fig:difference}, the difference of outputs exhibit distinct patterns across various models. In Open Sora, the pattern forms a 'U' shape, whereas in Latte and OpenSora-Plan, it resembles a horizontally flipped 'L'. Additionally, OpenSora-Plan features multiple peaks because its scheduler samples certain timesteps twice. The noisy input across consecutive timesteps changes minimally and shows little correlation with the model output. In contrast, both the timestep embedding and the timestep embedding modulated noisy input demonstrate a strong correlation with the model output. Given that the timestep embedding modulated noisy inputs exhibits superior generation capabilities (\eg, in Open Sora) and effectively leverages the dynamics of input, we select it as the indicator to determine whether the model output at the current step is similar to that of the previous timestep.

\subsection{TeaCache}
As illustrated in Fig.~\ref{fig:difference}, adjacent timesteps conduct redundant computations where model outputs exhibit minimal change. To minimize these redundancies and accelerate inference, we propose the Timestep Embedding Aware Cache (TeaCache). Rather than computing new outputs at each timestep, we reuse cached outputs from previous timesteps. Our caching technique can be applied to nearly all recent diffusion models based on Transformers.

\textbf{Naive Caching Strategy.} To determine whether to reuse the cached model output from a previous timestep, we employ the accumulated relative L1 distance as an indicator. 
\begin{equation}
\sum_{t=t_a}^{t_b-1} \text{L1}_{\text{rel}}(\mathbf{F}, t) \leq \delta < \sum_{t=t_a}^{t_b} \text{L1}_{\text{rel}}(\mathbf{F}, t)
\end{equation}
where $\text{L1}_{\text{rel}}$ is defined in Eq.~\ref{eq:difference_measurement}. $F$ can be timestep embedding or timestep embedding modulated noisy inputs and $\delta$ is the caching threshold. Specifically, after computing the model output at timestep $t_a$ and caching it, we accumulate the relative L1 distance $\sum_{t=t_a}^{t_b-1} \text{L1}_{\text{rel}}(\mathbf{F}, t)$ for subsequent timesteps. If, at timestep $t_b$ ($>t_a$), $\sum_{t=t_a}^{t_b-1} \text{L1}_{\text{rel}}(\mathbf{F}, t)$ is less than the caching threshold $\delta$, we reuse the cached model output; otherwise, we compute the new model output and set the accumulated relative L1 distance to zero.
A smaller threshold $\delta$ results in more frequent refreshing of cached outputs, while a larger threshold speeds up visual generation but may adversely affect image appearance. The threshold $\delta$ should be chosen to enhance inference speed without compromising visual quality.

\textbf{Rescaled Caching Strategy.} Although timestep embedding modulated noisy inputs exhibit a strong correlation with model outputs, the differences in consecutive timesteps are inconsistent. Directly using the difference of timestep embedding-modulated noisy input to estimate model output difference leads to a scaling bias. Such bias may cause suboptimal timestep selection. Considering that these differences are scalars, we apply simple polynomial fitting to rescale them to reduce the bias. The polynomial fitting is then performed between model inputs (timestep embedding modulated noisy inputs) and outputs, which is formulated as 
\begin{equation}
    y = f(x) = a_0 + a_1 x + a_2 x^2 + \cdots + a_n x^n,
\end{equation}
where $y$ represents estimated difference of model output and $x$ signifies the difference of timestep embedding-modulated noisy inputs. This can be efficiently solved using the \textit{poly1d} function from the numpy package.
With polynomial fitting, the rescaled difference in timestep embedding-modulated noisy inputs better estimates model output difference, as shown in Fig.~\ref{fig:fitting}. The final caching indicator is formulated as
\begin{equation}
\label{eq:thresh}
\sum_{t=t_a}^{t_b-1} f(\text{L1}_{\text{rel}}(\mathbf{F}, t)) \leq \delta < \sum_{t=t_a}^{t_b} f(\text{L1}_{\text{rel}}(\mathbf{F}, t))
\end{equation}

\begin{table*}[]
    \centering
    \caption{Quantitative evaluation of inference efficiency and visual quality in video generation models. TeaCache consistently achieves superior efficiency and better visual quality across different base models, sampling schedulers, video resolutions, and lengths.
    }
    \label{tab: main}
    \begin{tabular}{c|ccc|cccc}
        \toprule
        \multirow{2}{*}{\textbf{Method}} & \multicolumn{3}{c|}{\textbf{Efficiency}} & \multicolumn{4}{c}{\textbf{Visual Quality}} \\ \cline{2-8}
         & \textbf{FLOPs (P) $\downarrow$} & \textbf{Speedup $\uparrow$} & \textbf{Latency (s) $\downarrow$} & \textbf{VBench $\uparrow$} & \textbf{LPIPS $\downarrow$} & \textbf{SSIM $\uparrow$} & \textbf{PSNR $\uparrow$} \\

         \hline
        \hline
        \multicolumn{8}{c}{\textbf{Latte} (16 frames, 512$\times$512)} \\
        \hline
        \rowcolor[gray]{0.9} Latte $(T = 50)$ & 3.36 & 1$\times$ & 26.90 & 77.40\% & - & - & - \\
        $\Delta$-DiT~\cite{chen2024delta} & 3.36 & 1.02$\times$ & - & 52.00\% & 0.8513 & 0.1078 & 8.65 \\
        T-GATE~\cite{zhang2024cross} & 2.99 & 1.13$\times$ & - & 75.42\% & 0.2612 & 0.6927 & 19.55 \\
        PAB-slow~\cite{zhao2024real} & 2.70 & 1.21$\times$ & 22.16  &76.32\% & 0.2669 & 0.7014 &  19.71 \\
        PAB-fast~\cite{zhao2024real} & 2.52 & 1.34$\times$ & 19.98 & 73.13\% & 0.3903 & 0.6421 &  17.16 \\
        \hline
        TeaCache-slow & 1.86  &1.86 $\times$ & 14.46 & \textbf{77.40}\% & \textbf{0.1901} & \textbf{0.7786} &\textbf{22.09}  \\
        TeaCache-fast & \textbf{1.12}  &\textbf{3.28} $\times$ & \textbf{8.20} & 76.69\% & 0.3133 & 0.6678 &18.62  \\

        \hline
        \hline
        \multicolumn{8}{c}{\textbf{Open-Sora 1.2} (51 frames, 480P)} \\
        \hline
        \rowcolor[gray]{0.9} Open-Sora 1.2 $(T = 30)$ & 3.15 & 1$\times$ & 44.56 & 79.22\% & - & - & - \\
        $\Delta$-DiT~\cite{chen2024delta} & 3.09 & 1.03$\times$ & - & 78.21\% & 0.5692 & 0.4811 & 11.91 \\
        T-GATE~\cite{zhang2024cross} & 2.75 & 1.19$\times$ & - & 77.61\% & 0.3495 & 0.6760 & 15.50 \\
        PAB-slow~\cite{zhao2024real} & 2.55 & 1.33$\times$ & 33.40  & 77.64\% & 0.1471 & 0.8405 &  \textbf{24.50} \\
        PAB-fast~\cite{zhao2024real} & 2.50 & 1.40$\times$ & 31.85 & 76.95\% & 0.1743 & 0.8220 &  23.58 \\
        \hline
        TeaCache-slow & 2.40 & 1.55$\times$ & 28.78 & \textbf{79.28}\% & \textbf{0.1316} & \textbf{0.8415} &23.62  \\
        TeaCache-fast & \textbf{1.64} & \textbf{2.25}$\times$ & \textbf{19.84} & 78.48\% & 0.2511 & 0.7477 & 19.10 \\

        \hline
        \hline
        \multicolumn{8}{c}{\textbf{OpenSora-Plan} (65 frames, 512$\times$512)} \\
        \hline
        \rowcolor[gray]{0.9} OpenSora-Plan $(T = 150)$ & 11.75 & 1$\times$ &99.65  & 80.39\% & - & - & - \\
        $\Delta$-DiT~\cite{chen2024delta} & 11.74 & 1.01$\times$ & - & 77.55\% & 0.5388 & 0.3736 & 13.85 \\
        T-GATE~\cite{zhang2024cross} & 2.75 & 1.18$\times$ & - & 80.15\% & 0.3066 & 0.6219 & 18.32 \\
        PAB-slow~\cite{zhao2024real} & 8.69 & 1.36$\times$ & 73.41  &80.30\% & 0.3059 & 0.6550 &  18.80 \\
        PAB-fast~\cite{zhao2024real} & 8.35 & 1.56$\times$ & 65.38 & 71.81\% & 0.5499 & 0.4717 &  15.47 \\
        \hline
        TeaCache-slow & 3.13 & 4.41$\times$ & 22.62 & \textbf{80.32}\% & \textbf{0.2145} & \textbf{0.7414} & \textbf{21.02}  \\
        TeaCache-fast & \textbf{2.06}  & \textbf{6.83}$\times$ & \textbf{14.60} & 79.72\% & 0.3155 & 0.6589 & 18.95 \\

        \bottomrule
    \end{tabular}
\end{table*}

\subsection{Discussion}
\textbf{Caching Mechanism \textit{v.s.} Reducing Timesteps.} 
Assume that both of the caching mechanism and reducing timesteps stregegies reduces half of the timesteps. The differences between them can be concluded in three aspects: 
(1) Timesteps. Our caching strategy dynamically selects timesteps with large difference for caching and reusing in the following timesteps, whereas the strategy of reducing timesteps is conducted uniformly, lacking awareness of dynamic differences among different timesteps. 
(2) Model output. In our caching strategy, only the residual signal (\ie, Output minus Input) in the diffusion transformer is cached, therefore the model output at the next timestep is updated. In contrast, reducing timesteps can be considered as keeping the output constant in the next timestep.
(3) Parameter $\alpha_t$. Reducing timesteps results in a coarser-grained $\alpha_t$, which suffers from deteriorated visual quality. In comparison, TeaCache is able to maintain the visual quality, as illustrated in Fig.~\ref{fig:few timestep}.

\section{Experiment}

\begin{figure*}
  \centering
  \includegraphics[width=0.98\linewidth]{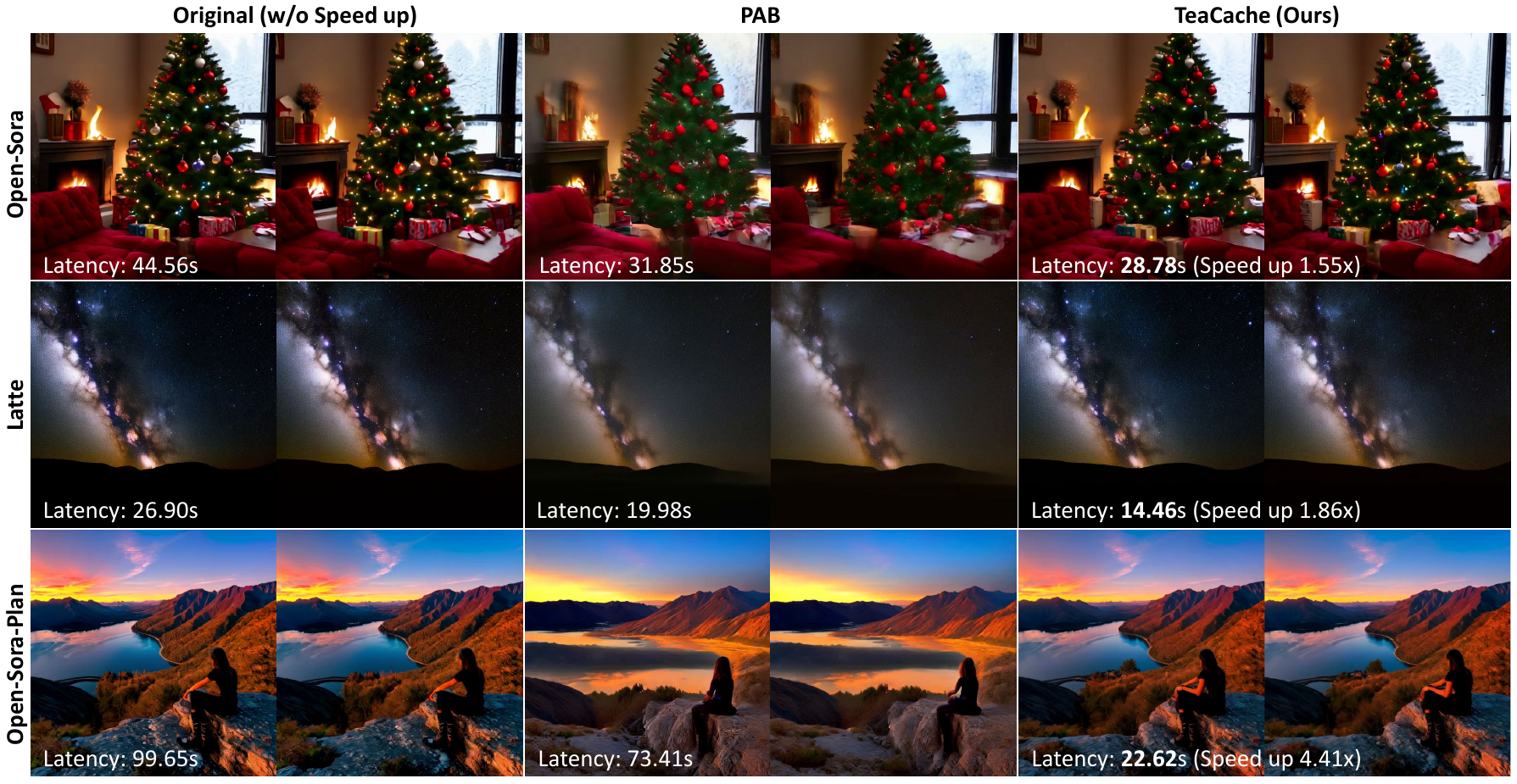}
  \caption{
  Comparison of visual quality and efficiency (denoted by latency) with the competing method. TeaCache outperforms PAB~\cite{zhao2024real} in both visual quality and efficiency. Latency is evaluated on a single A800 GPU. Video generation specifications: Open-Sora~\cite{Open-Sora} (51 frames, 480p), Latte~\cite{ma2024latte} (16 frames, 512$\times$512), Open-Sora-Plan~\cite{Open-Sora-Plan} (65 frames , 512$\times$512). Best-viewed with zoom-in.
  }
  \label{fig:show}
\end{figure*}

\subsection{Settings}
\textbf{Base Models and Compared Methods.} To demonstrate the effectiveness of our method, we apply our acceleration technique to various video, such as Open-Sora 1.2 ~\cite{Open-Sora}, Open-Sora-Plan~\cite{Open-Sora-Plan} and Latte~\cite{ma2024latte}. We compare our base models with recent efficient video synthesis techniques, including PAB~\cite{zhao2024real}, T-GATE~\cite{zhang2024cross} and $\Delta$-DiT~\cite{chen2024delta}, to highlight the advantages of our approach. Notably, $\Delta$-DiT and T-GATE are originally designed as an acceleration method for image synthesis. PAB adapted them for video synthesis to facilitate comparison. 

\textbf{Evaluation Metrics.} To assess the performance of video synthesis acceleration methods, we focus on two primary aspects: inference efficiency and visual quality. For evaluating inference efficiency, we use Floating Point Operations (FLOPs) and inference latency as metrics. 
Following PAB~\cite{zhao2024real},  we employ VBench~\cite{huang2024vbench}, LPIPS~\cite{zhang2018unreasonable}, PSNR, and SSIM for visual quality evaluation.

\textbf{Implementation Detail} All experiments are carried out on the NVIDIA A800 80GB GPUs with Pytorch. We enable FlashAttention~\cite{dao2022flashattention} by default for all experiments.  To obtain robust polynomial fitting, we sample 70 prompts from T2V-CompBench~\cite{sun2024t2v} to generate videos, assessing seven desired attributes of generated videos. 10 prompts are sampled for each attributes.
$\delta$ in Eq.~\ref{eq:thresh} is 0.1 for TeaCache-slow and 0.2 for TeaCache-fast.

\begin{figure*}
    \centering
    \begin{minipage}{\textwidth}
    \centering
    \begin{subfigure}{0.24\textwidth}
        \centering
        \includegraphics[width=\textwidth]{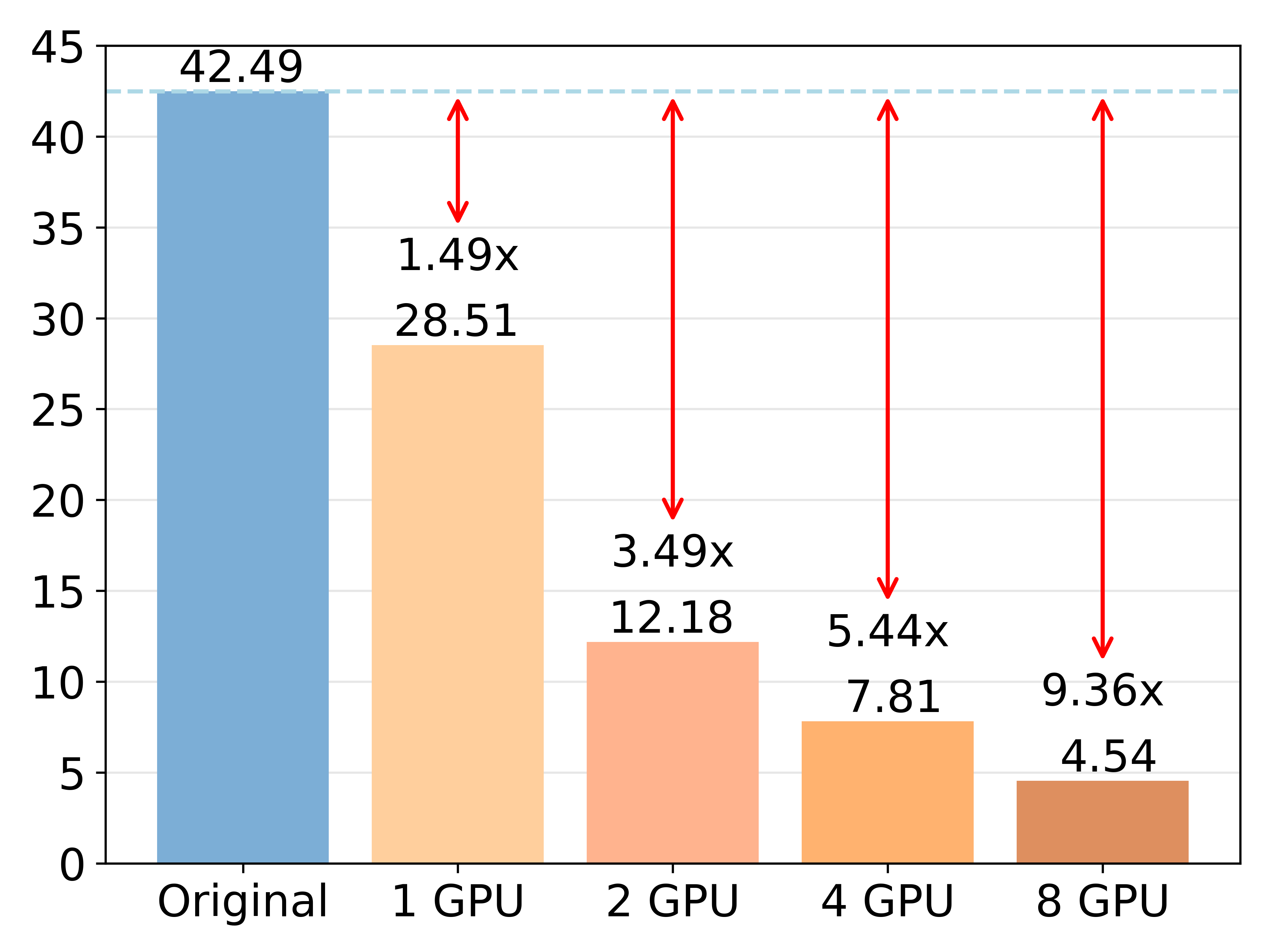} 
        \caption{480P, 48 frames}
    \end{subfigure}
    \hfill
    \begin{subfigure}{0.24\textwidth}
        \centering
        \includegraphics[width=\textwidth]{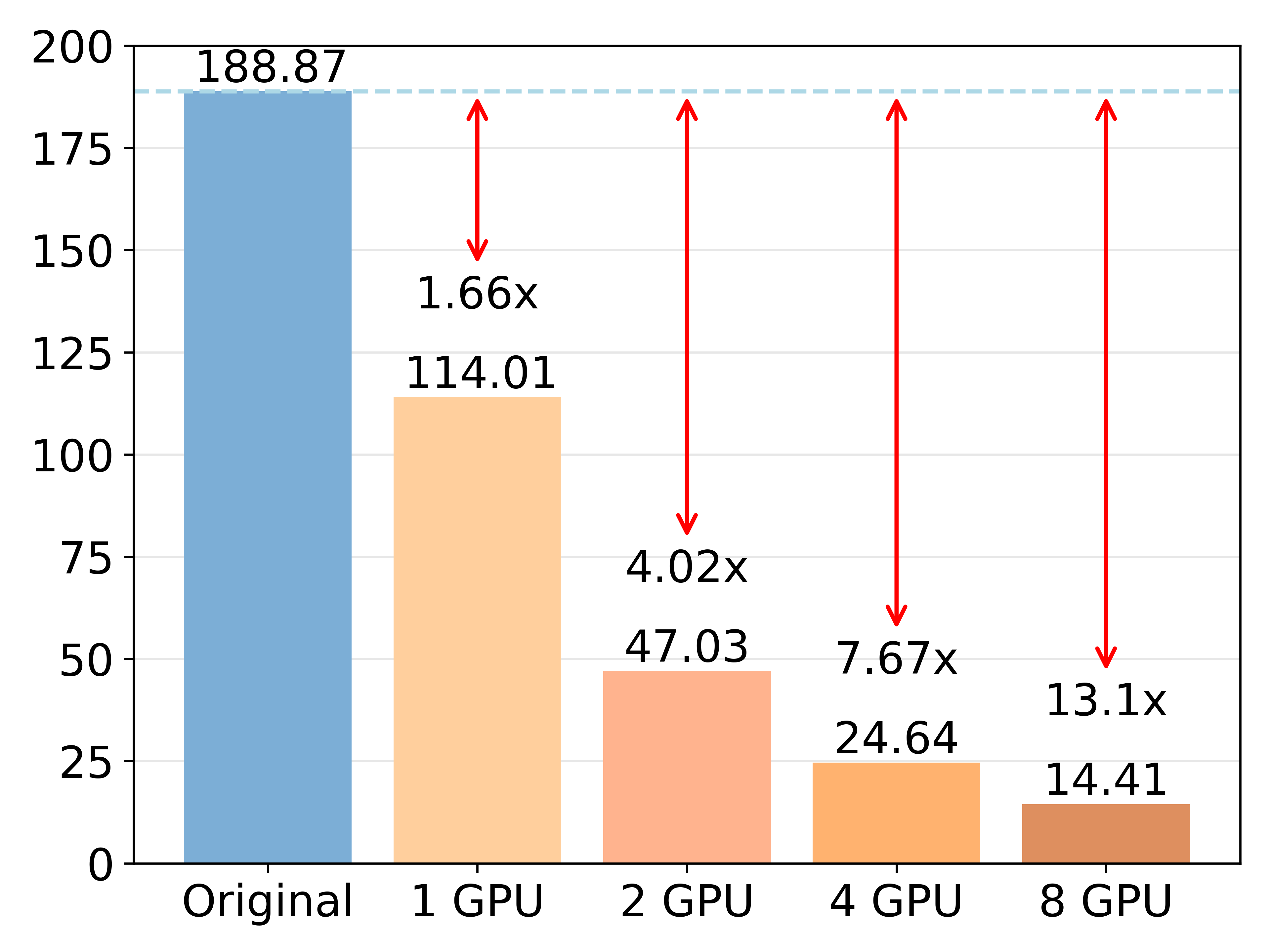} 
        \caption{480P, 192 frames}
    \end{subfigure}
    \hfill
    \begin{subfigure}{0.24\textwidth}
        \centering
        \includegraphics[width=\textwidth]{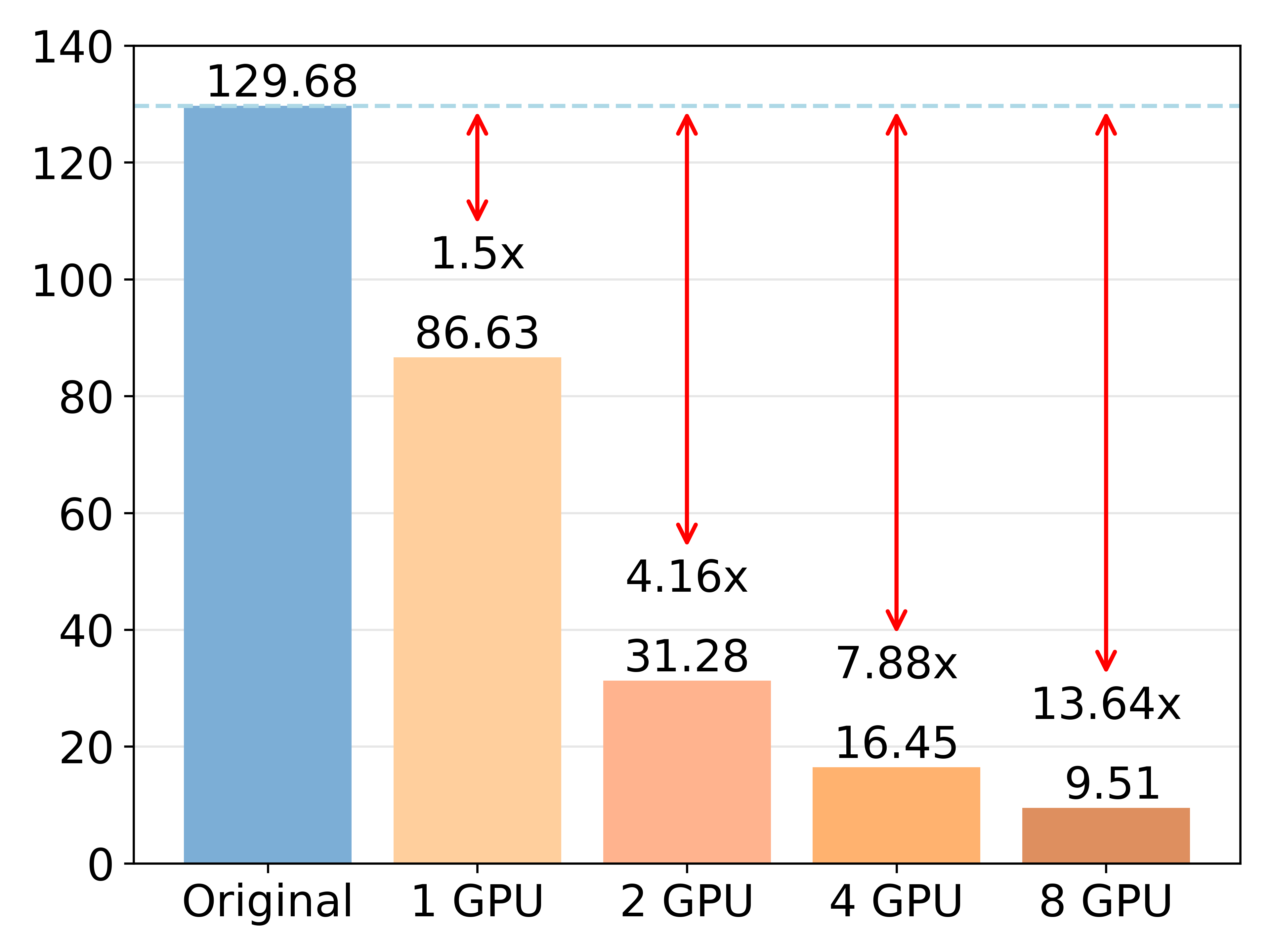}
        \caption{360P, 240 frames}
    \end{subfigure}
    \hfill
    \begin{subfigure}{0.24\textwidth}
        \centering
        \includegraphics[width=\textwidth]{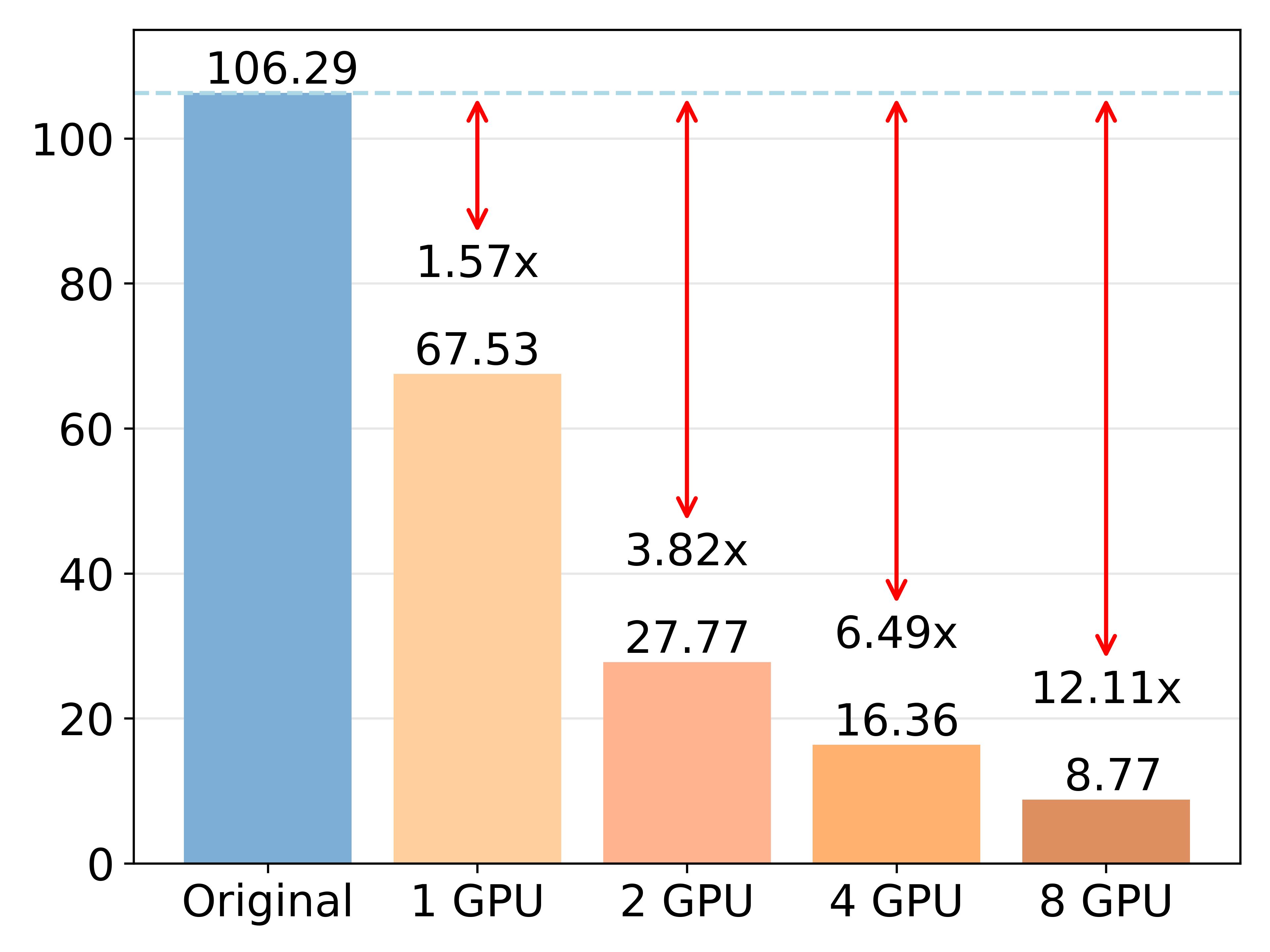}
        \caption{720P, 48 frames}
    \end{subfigure}

    \end{minipage}

    \caption{Inference efficiency of TeaCache at different video lengths and resolutions.}
    \label{fig:multi resolution}
\end{figure*}

\subsection{Main Results}

\textbf{Quantitative Comparison.} Tab.~\ref{tab: main} presents a quantitative evaluation of efficiency and visual quality using the VBench benchmark ~\cite{huang2024vbench}. We examine two variants of TeaCache: a slow variant and a fast variant with greater speedup. Compared to other training-free acceleration methods, TeaCache consistently achieves superior efficiency and better visual quality across different base models, sampling schedulers, video resolutions, and lengths. In evaluating the Latte~\cite{ma2024latte} baseline, the TeaCache-slow model demonstrates superior performance across all visual quality metrics, achieving a 1.86× speedup compared to PAB~\cite{zhao2024real}, which provides a 1.34× speedup. TeaCache-fast achieves the highest acceleration at 3.28×, albeit with a slight reduction in visual quality. With the OpenSora~\cite{Open-Sora} baseline, we obtain the optimal speedup of 2.25× as compared to the previous 1.40×, and the highest overall quality with a speedup of 1.55×. Additionally, using Open-Sora-Plan~\cite{Open-Sora-Plan}, TeaCache achieves the highest speedup of 6.83×, surpassing the previously best 1.49× offered by PAB, while also delivering the highest quality at a speedup of 4.41×.

\textbf{Visualization.} Fig.~\ref{fig:show} compares the videos generated by TeaCache against those by the original model and  PAB. The results demonstrate that TeaCache outperforms PAB in visual quality with lower latency.

\subsection{Ablation Studies}

\textbf{Scaling to multiple GPUs.} Aligned with previous research employing Dynamic Sequence Parallelism (DSP)~\cite{zhao2024real} for supporting high-resolution long-video generation across multiple GPUs, we assess the performance of TeaCache in these scenarios. 
The results of this study are presented in Tab.~\ref{tab: multiple GPU}. We utilize Open-Sora~\cite{Open-Sora} and Open-Sora-Plan~\cite{Open-Sora-Plan} as baselines and compare them against the prior method PAB~\cite{zhao2024real} regarding latency measurements on A800 GPUs. As the number of GPUs increases, TeaCache consistently improves inference speed across various base models and outperforms PAB.

\textbf{Performance at different Length and Resolution.} To assess the effectiveness of our method in accelerating sampling for videos with varying sizes, we perform tests across different video lengths and resolutions. The results, presented in Fig.~\ref{fig:multi resolution}, demonstrate that our method sustains consistent acceleration performance, even with increases in video resolution and frame count. This consistency highlights the method's potential to accelerate sampling processes for longer and higher-resolution videos, meeting practical demands.

\textbf{Quality-Efficiency trade-off.} In Fig.~\ref{fig:shot}, we compare the quality-latency trade-off of TeaCache with PAB~\cite{zhao2024real}. The thresholds $\delta$ in Eq.~\ref{eq:thresh} are 0.1, 0.15, 0.2 and 0.25. Our analysis reveals that TeaCache achieves significantly higher reduction rates, indicated by lower absolute latency, compared to PAB. Additionally, across a wide range of latency configurations, TeaCache consistently outperforms PAB on all quality metrics. This is particularly evident in the reference-free metric VBench score~\cite{huang2024vbench}, which aligns more closely with human preferences. Although there is a decline in reference-based scores such as PSNR and SSIM at extreme reduction rates, qualitative results suggest that the outputs remain satisfactory, despite not perfectly matching the reference.

\textbf{Choice of Indicator.} When determining the caching schedule, we evaluate various indicators to estimate the differences in model outputs across consecutive timesteps, including timestep embedding and timestep embedding-modulated noisy input. As illustrated in Fig.~\ref{fig:difference}, the timestep embedding-modulated noisy input demonstrates a stronger correlation with model output compared to the timestep embedding, particularly in the OpenSora. Moreover, the selection of timesteps by the timestep embedding-modulated noisy input adapts dynamically to different prompts, whereas the timestep embedding selects the same timesteps for all prompts. This observation is validated by the results presented in Tab.~\ref{tab:indicator}, where the timestep embedding-modulated noisy input consistently surpasses the timestep embedding across various models.

\textbf{Effect of Rescaling.} Tab.\ref{tab:fitting} illustrates the impact of rescaling. A first-order polynomial fitting outperforms the original data by 0.24\% under Vbench score metric, as well as in LPIPS, SSIM, and PSNR metrics. Performance gains tend to saturate with a fourth-order polynomial fitting.

\begin{table}[]
\scriptsize
\centering
    \caption{Ablation study of caching indicator. `Timestep': timestep embedding. `Input': timestep embedding-modulated noisy input.}
    \label{tab:indicator}
\begin{tabular}{c|cccc}
\toprule
\textbf{Indicator}             &\textbf{VBench $\uparrow$} & \textbf{LPIPS $\downarrow$} & \textbf{SSIM $\uparrow$} & \textbf{PSNR $\uparrow$} \\
\hline
\rowcolor[gray]{0.9}OpenSora              & 79.22\%          &  -         &  -         &  -    \\
Timestep    & 77.01\% & 0.3425          & 0.6934          & 15.86     \\
Input & \textbf{78.21}\%  & \textbf{0.2549}  & \textbf{0.7457}  & \textbf{19.05}    \\
\hline
\rowcolor[gray]{0.9}Latte              & 77.40\%          &  -         &  -         &  -    \\
Timestep    &77.05\%  & 0.2653          &   0.7073        &  19.76    \\
Input & \textbf{77.17}\% & \textbf{0.2558} & \textbf{0.7164} & \textbf{20.00}  \\
\bottomrule
\end{tabular}
\end{table}

\vspace{-0.5em}

\begin{table}[]
\scriptsize
\centering
    \caption{Ablation study of polynomial fitting. Rescaling with polynomial fitting outperforms original data. Higher-order fitting obtains better performance and saturates in 4-order fitting. }
    \label{tab:fitting}
\begin{tabular}{c|cccc}
\toprule
\textbf{Order}             &\textbf{VBench $\uparrow$} & \textbf{LPIPS $\downarrow$} & \textbf{SSIM $\uparrow$} & \textbf{PSNR $\uparrow$} \\
\hline
\rowcolor[gray]{0.9}OpenSora              & 79.22\%          &  -         &  -         &  -    \\
Original    & 78.21\%  & 0.2549  & 0.7457  & 19.05     \\
1-order &78.45\%  &0.2517  &\textbf{0.7478}  &19.10   \\
2-order &78.48\%  &0.2513  &0.7477  &19.09   \\
4-order &\textbf{78.48}\% & \textbf{0.2511} & 0.7477 & \textbf{19.10}   \\
\bottomrule
\end{tabular}
\end{table}

\begin{table}[]
\scriptsize
\centering
\caption{Inference efficiency and visual quality when scaling to multiple GPUs with Dynamic Sequence Parallelism (DSP).}
\label{tab: multiple GPU}
\scalebox{0.92}{
\begin{tabular}{l|c|c|c|c}
\toprule
\textbf{Method} & 1 × A800 & 2 × A800 & 4 × A800 & 8 × A800 \\ 
\hline
\multicolumn{5}{c}{\textbf{Open-Sora (192 frames, 480P)}} \\ \hline
Baseline &  188.87(1×) &  72.86(2.59×) &  39.26(4.81×) &  22.18(8.52×) \\ \hline
PAB &  142.23(1.33×) &  53.74(3.51×) &  29.19(6.47×) &  16.88(11.19×) \\ \hline
\rowcolor{gray!30} TeaCache &  114.01(1.66×) &  47.03(4.02×) &  24.64(7.67×) &  14.41(13.10×) \\ \hline
\multicolumn{5}{c}{\textbf{Open-Sora-Plan (221 frames, 512×512)}} \\ \hline
Baseline &  324.41(1×) &  166.94(1.94×) &  88.18(3.68×) &  47.79(6.79×) \\ \hline
PAB &  207.70(1.56×) &  110.06(2.95×) &  58.07(5.59×) &  31.92(10.16×) \\ \hline
\rowcolor{gray!30} TeaCache &  48.22(6.73×) &  26.99(12.02×) &  15.91(20.39×) & 10.13 (32.02×) \\ 
\bottomrule
\end{tabular}
}
\end{table}

\section{Conclusion}
In this study, we introduce \textbf{TeaCache}, a novel, training-free approach designed to significantly accelerate video synthesis inference while maintaining high-quality output. We analyze the correlation between model input and output,  observing that similarity of timsetep embedding modulated noisy input in consecutive timesteps shows strong correlation with similarity of model output. We propose to utilize similarity of timsetep embedding modulated noisy input as an indicator of output similarity, allowing for dynamic caching of model outputs. Further, we propose a rescaling strategy to refine the estimation of model output similarity,  optimizing the selection of timestep. Extensive experiments demonstrate TeaCache's robust performance in terms of both efficiency and visual quality across diverse video generation models and image generation models, sampling schedules, video lengths, and resolutions, underscoring its potential for real-world applications.

\noindent\textbf{Acknowledgment.} This work was supported in part by the National Natural Science Foundation
of China (NSFC) underGrants 62472402,  the Fundamental Research Funds for the Central Universities, National Natural Science Foundation of China (NSFC) under Grant 62225208 and CAS Project for CAS Project for Young Scientists in Basic Research under Grant No.YSBR-117 and Alibaba Research Intern Program.

{
    \small
    \bibliographystyle{ieeenat_fullname}
    \bibliography{main}
}

\end{document}